\documentclass[a4paper,fleqn]{cas-sc}

\usepackage[natbib=true, style=numeric]{biblatex}
\usepackage{tabularx}
\usepackage{graphicx}     
\usepackage{subcaption}   
\usepackage{caption}      
\usepackage{pifont}
\usepackage{multirow}
\usepackage{graphicx}
\usepackage{colortbl}
\usepackage{booktabs}
\usepackage{xcolor}
\usepackage{adjustbox}
\usepackage{cleveref}

\newcommand{\cmark}{\checkmark} 
\newcommand{\xmark}{\ding{55}} 

\begin{document}
\makeatletter
\def\printFirstPageNotes{}
\makeatother

\let\WriteBookmarks\relax
\def\floatpagepagefraction{1}
\def\textpagefraction{.001}

\shorttitle{}

\shortauthors{Y. Borhani et~al.}

\title [mode = title]{PoseDriver: A Unified Approach to Multi-Category Skeleton Detection for Autonomous Driving}                      



%

\author[1]{Yasamin Borhani}
\author[1]{Taylor Mordan}
\author[1]{Yihan Wang}
\author[1]{Reyhaneh Hosseininejad}
\author[1]{Javad Khoramdel}
\author[1]{Alexandre Alahi}

\affiliation[1]{Ecole Polytechnique Federale de Lausanne (EPFL)}

\begin{abstract}
Object skeletons offer a concise representation of structural information, capturing essential aspects of posture and orientation that are crucial for autonomous driving applications. However, a unified architecture that simultaneously handles multiple instances and categories using only the input image remains elusive. In this paper, we introduce PoseDriver, a unified framework for bottom-up multi-category skeleton detection tailored to common objects in driving scenarios. We model each category as a distinct task to systematically address the challenges of multi-task learning. Specifically, we propose a novel approach for lane detection based on skeleton representations, achieving state-of-the-art performance on the OpenLane dataset. Moreover, we present a new dataset for bicycle skeleton detection and assess the transferability of our framework to novel categories. Experimental results validate the effectiveness of the proposed approach.

\end{abstract}









\maketitle

\section{Introduction}
The imperative to detect and understand dynamic objects in autonomous driving scenes is well-established within the literature. While early efforts focused on identifying the presence of objects like pedestrians, often using bounding boxes \cite{alahi2014robust}, it became evident that such representations are insufficient for nuanced interaction. For instance, anticipating a pedestrian's intent can hinge on subtle cues like head orientation, a detail that requires more granular information than a bounding box can provide \cite{kalatian2022context}. This need for detail extends to interactions between vehicles, where advanced trajectory and motion planning frameworks increasingly presume access to rich predictive data about surrounding agents' behaviors and future paths \cite{zhang2023predictive, dong2024enhanced}. A persistent challenge, however, is that such sophisticated planning models often rely on data that is difficult to capture with a single, efficient perception system. This creates a critical gap between the requirements of high-level decision-making algorithms and the practical capabilities of real-time environmental perception. To bridge this gap, a more descriptive and versatile representation is required.

Skeleton representations provide a compact and expressive way to depict objects by explicitly outlining their structural components. Unlike segmentation masks and bounding boxes, which provide coarse outlines or holistic object representations, skeletons capture the spatial configuration and structural details of objects. Such representations are particularly useful in autonomous driving scenarios, where accurate object understanding enhances safety and decision making \cite{gesnouin2020predicting, mordan2021detecting}. Their detailed yet efficient nature benefits downstream tasks, such as depth estimation \cite{bertoni2019monoloco} and human action recognition \cite{xiong2022simple, zhu2023motionbert, rajasegaran2023benefits, abdelfattah2024sjepa}, while also improving model interpretability by focusing on key structural features that can aid in explaining prediction outcomes.

In this paper, we present a unified architecture for representing key scene elements in autonomous driving. Our approach captures both dynamic objects, including cars, animals, pedestrians, and bicycles, and static elements such as lanes. Using skeleton-based representations, we effectively model the structural details of both deformable objects (e.g., animals and pedestrians) and rigid ones (e.g., cars and bicycles). Additionally, we extend this representation to lanes, which traditionally lack skeleton-based annotations. Unlike dynamic objects, which have well-defined keypoints, lanes present a unique challenge due to the absence of clearly identifiable markers. Within the scope of emerging technologies in transportation, various studies have advanced object detection and tracking for dynamic road users \cite{zhao2019detection, fonod2025advanced}, while others have explored multi-task learning paradigms to enhance comprehensive scene understanding in autonomous systems \cite{munir2025context, kalatian2022context, cantarella2022multi}.

\begin{figure}
    \centering
    \includegraphics[width=0.8\textwidth]{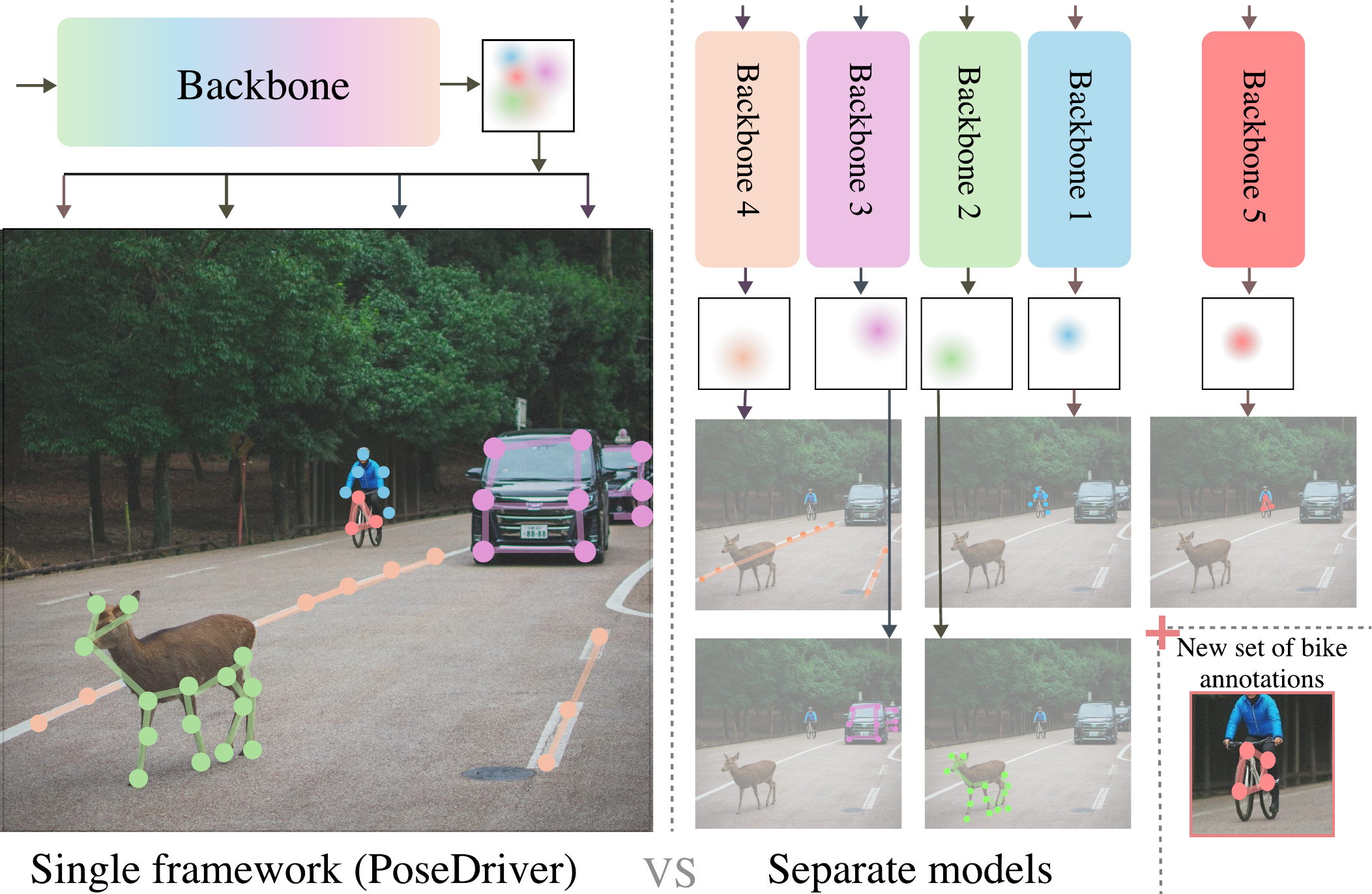} 
    \caption{
    Skeleton detection provides a detailed but light representation of the environment for autonomous driving. Our purpose is to jointly detect skeletons on dynamic road users (cars, humans, and animals) as well as static road configuration (lanes) to get a comprehensive representation of the environment around the car, enabling a better understanding and safer navigation.}
    \label{fig:pull}
\end{figure}

Although single-category keypoint and skeleton detection (e.g., for pedestrians \cite{cao2017realtime, xu2022vitpose, kreiss2019pifpaf, kreiss2021openpifpaf}, animals \cite{ye2022superanimal, cao2019cross, kreiss2021openpifpaf}, and cars \cite{wei2016convolutional,kreiss2021openpifpaf}) is well studied, multicategory skeleton detection remains less explored. Recent models using text or image prompts enable category-agnostic keypoint detection \cite{shi2023matching, hirschorn2023pose, xu2022pose, nguyen2024escape, yang2025x}, but often detect only one instance per image and rely on object detectors, annotated prompts, multiple inference passes, and extra encoders. X-Pose \cite{yang2025x} removes the single-instance limitation but still requires prompt encoders, adding complexity. These models generalize well to familiar categories (e.g., transferring keypoints between similar objects) but struggle with novel ones like bicycles or lanes, often requiring costly re-training.

We introduce PoseDriver, a unified framework for bottom-up joint detection of skeletons across five key categories in autonomous driving: cars, pedestrians, bicycles, animals, and lanes. Our experiments show that PoseDriver improves multi-task keypoint detection performance across various backbone architectures. To mitigate domain shift issues across our diverse datasets, we critically examine the role of normalization layers in backbone networks. Specifically, we assess the limitations of Batch Normalization in multi-domain training and adopt architectures like ConvNeXt and Swin that are designed without it. To address performance degradation from batch normalization, we exclude batch normalization layers, following prior studies. We also present a new COCO bicycle keypoint detection dataset and a novel lane detection approach using evenly distributed keypoints, preserving lane structure and topological continuity. Our results demonstrate that multi-task training weights can be effectively transferred to unseen categories, improving performance without additional computational cost compared to single-task training.

The main contributions of this paper are as follows: 
\begin{itemize}  
    \item \textbf{Bottom-up Multi-Category Skeleton Detection:} We identify and address key challenges in multi-category skeleton detection by proposing a unified bottom-up architecture that jointly detects cars, bicycles, pedestrians, animals, and lanes. Our experiments show that, aside from a slight drop in pedestrian detection, the framework matches or outperforms single-category models.

    \item \textbf{Skeleton-Based Lane Detection:} We propose a novel lane detection method that represents lanes as a fixed set of keypoints, capturing both global structure and local continuity. Our evaluations of the benchmark datasets confirm its competitive performance.

    \item \textbf{Bicycle Skeleton Annotation and Multi-Task Transfer Learning
:} We introduce a novel annotation set of bicycle skeletons on a subset of the MS COCO dataset, thereby extending our multi-category skeleton detection framework to a previously unaddressed category. Experimental results indicate that models pre-trained in a multi-task setting, which leverages shared representations across various object categories, outperform those trained solely on bicycle skeleton detection.
\color{black}
\end{itemize}

\section{Related Works}

\label{sec:related_works}

\subsection{Skeleton Detection}
The main objective of skeleton detection is to localize and link keypoints on various objects, which can be challenging in crowded scenes, with varying scales, occlusions, and rare poses. Methods are generally classified as top-down or bottom-up. Top-down approaches, like Mask R-CNN \cite{he2017mask} and ViTPose \cite{xu2022vitpose}, first, detect bounding boxes before applying keypoint detection within each box. ViTPose uses transformer-based self-attention for long-range dependencies, while Polarized Self-Attention (PSA) \cite{liu2021polarized} combines global and local detail for pixel-wise precision. In contrast, bottom-up methods detect all keypoints in an image first and then associate them with each object. Techniques such as Part Affinity Fields (PAFs) \cite{kreiss2021openpifpaf}, Associative Embedding \cite{newell2017associative}, PersonLab \cite{papandreou2018personlab}, and OpenPifPaf \cite{kreiss2019pifpaf, kreiss2021openpifpaf} follow this approach. OpenPifPaf, an open-source framework for real-time multi-person pose estimation, utilizes Composite Intensity Fields (CIF) and Composite Association Fields (CAF) for precise keypoint localization and association, directly detecting and linking keypoints across multiple people without requiring individual detections, making it effective in crowded or occluded scenes.

\begin{figure}[htbp]
  \centering
  \begin{subfigure}{0.48\linewidth}
    \includegraphics[width=\linewidth]{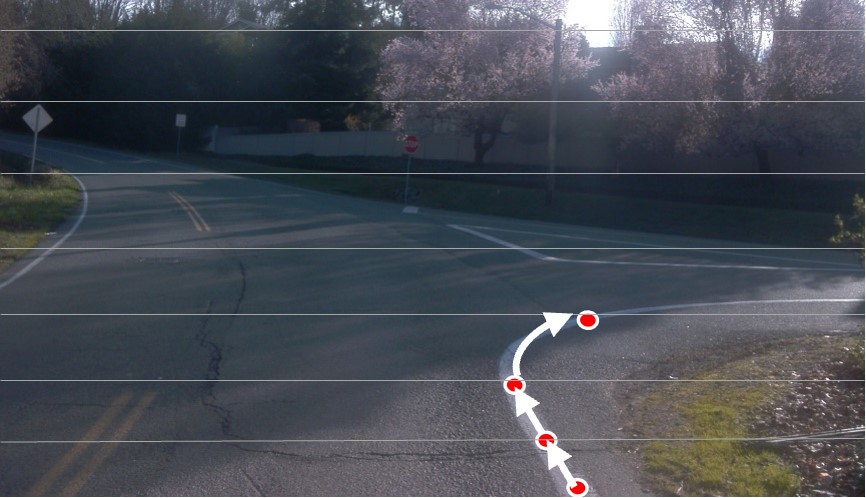}
    \caption{FoloLane \cite{qu2021focus}}
    \label{fig:fololane}
  \end{subfigure}
  \hfill
  \begin{subfigure}{0.48\linewidth}
    \includegraphics[width=\linewidth]{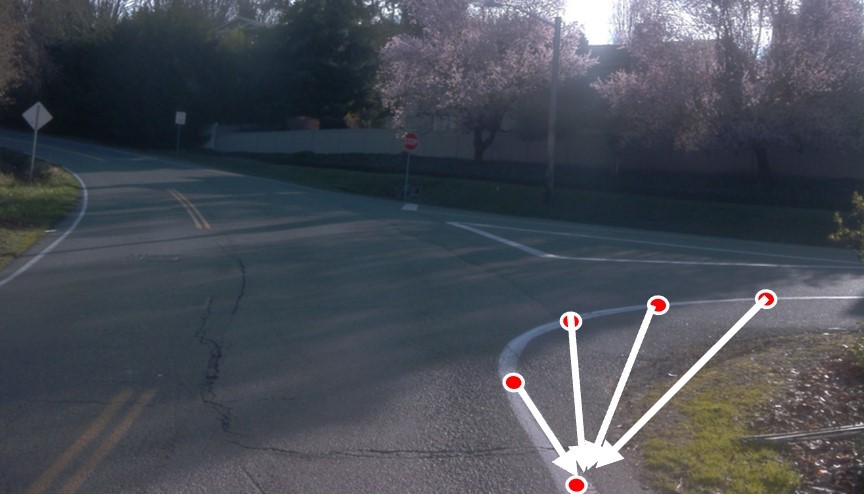}
    \caption{GANet \cite{wang2022keypoint}}
    \label{fig:ganet}
  \end{subfigure}

  \vspace{0.5em}

  \begin{subfigure}{0.48\linewidth}
    \includegraphics[width=\linewidth]{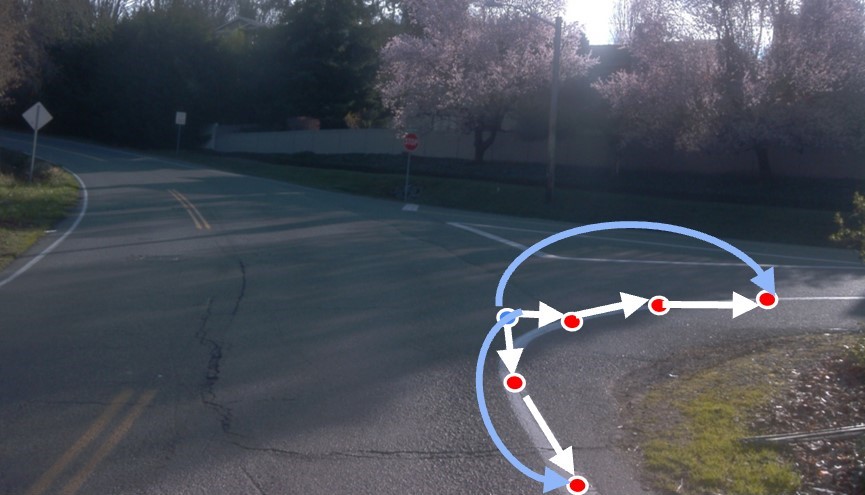}
    \caption{RCLane \cite{xu2022rclane}}
    \label{fig:RCLane}
  \end{subfigure}
  \hfill
  \begin{subfigure}{0.48\linewidth}
    \includegraphics[width=\linewidth]{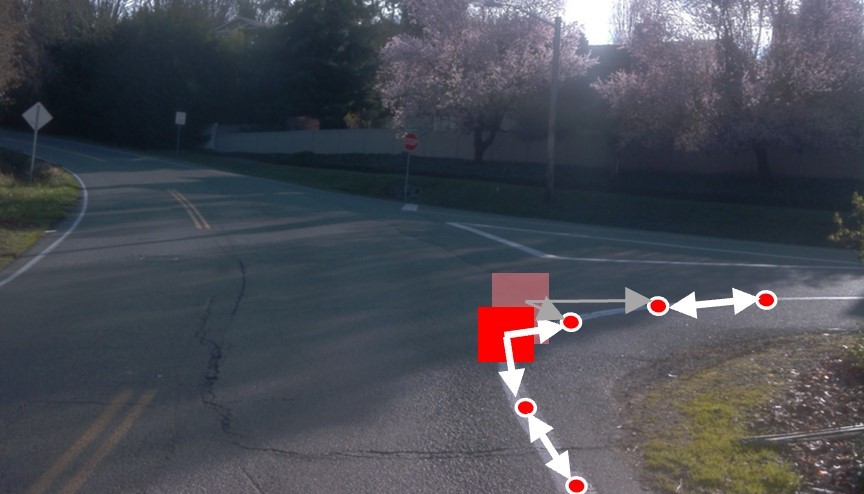}
    \caption{PoseDriver [Ours]}
    \label{fig:ours}
  \end{subfigure}

  \caption{Comparison of skeleton-based lane detection methods' association schemes. (a) FoloLane \cite{qu2021focus} follows a locally iterative manner. (b) GANet \cite{wang2022keypoint} regresses each keypoint to its starting point. (c) RCLane \cite{xu2022rclane} models lanes as relay chains. (d) PoseDriver jointly estimates keypoint location and association using intensity and association fields.}
  \label{fig:intro}
\end{figure}

\subsection{Lane Detection}
Deep learning based lane detection methods generally fall into four categories:

\textbf{Segmentation-based Methods.} These approaches treat lane detection as a pixel-wise classification task, generating lane masks. For example, SCNN \cite{pan2018spatial} and others \cite{hou2019learning, neven2018towards, zheng2021resa} use multi-class segmentation to distinguish lanes, but with high computational costs (e.g., SCNN achieves only 7.5 FPS), limiting real-time suitability.

\textbf{Anchor-based Methods.} These methods model lane detection as a top-down object detection task. Techniques like Line-CNN \cite{8624563} and Curve-NAS \cite{xu2020curvelane} employ line anchors, while LaneATT \cite{9412265} and CLRNet \cite{zheng2022clrnet} integrate attention and globally spaced anchors. However, pre-defined anchors struggle with complex lane shapes, like Y-shaped lanes.

\textbf{Curve-based Methods.} Curve-based methods use parametric models to represent lanes, such as polynomial regressions in PolyLaneNet \cite{9412265}, transformer-enhanced models \cite{liu2021end}, and Bézier curves \cite{feng2022rethinking}. While effective, these models lack flexibility in complex environments.

\textbf{Keypoint-based Methods.} Keypoint-based approaches, such as FoloLane \cite{qu2021focus} and RCLane \cite{xu2022rclane}, model lanes as a series of keypoints with associations. However, occlusions and unstable structures pose challenges. Our approach represents lanes with a fixed number of keypoints, balancing global structure and local correlations for consistent and efficient lane detection.

This line of work \cite{qu2021focus,wang2022keypoint,xu2022rclane} designs various association schemes to group the keypoints into different lanes, as illustrated in \Cref{fig:intro}. The key challenges for these methods are the global association of keypoints and preserving the continuity of lane topology, especially when lanes are partially occluded. This occurs often in situations such as crowded roads or due to the blur on the vehicle's front window. 


Unlike traditional methods that iteratively construct lane associations using local structures \cite{qu2021focus,xu2022rclane} or predefined reference points \cite{wang2022keypoint}, our approach detects entire lane skeletons in a single pass, with connectivity inherently encoded. We achieve a more streamlined and efficient process by eliminating the need for a step-by-step post-detection association. Additionally, our PAF-based framework outputs per-association heatmaps in parallel, significantly reducing post-processing overhead. Beyond methodological differences, our skeleton-based representation offers a fundamental advantage over unordered keypoint sets by naturally preserving global structural relationships, such as keypoint distances from the camera, enabling a richer and more coherent understanding of the scene.

\color{black}

\subsection{Category Agnostic Keypoint Detection}

Recent efforts toward unified networks for multi-category keypoint extraction include: a two-stage framework that separates keypoint matching from pose refinement \cite{shi2023matching}; a graph-based model leveraging keypoint relationships for generalized pose estimation \cite{hirschorn2023pose}; a transformer-based Keypoint Interaction Module (KIM) capturing interactions among keypoints and between support and query images \cite{xu2022pose}; and ESCAPE, a Bayesian framework that uses "super-keypoints" as abstract representations for adapting to new categories \cite{nguyen2024escape}. However, these approaches face limitations due to reliance on the small MP-100 dataset, dependence on query images with predefined keypoints, and the need for high similarity between query and support images.
The recently introduced X-Pose framework \cite{yang2025x} addresses these challenges by supporting multi-class keypoint detection with multi-modal prompts, enabling keypoint generation from both textual and visual queries and handling images with multiple objects. Although this framework enhances generalizability with a larger dataset, it requires both a CLIP image and text encoder, adding complexity and computational demands. Training also necessitates annotated visual data with predefined keypoints, and lane detection using keypoints remains unintegrated.

\section{Expanding Skeletons to Lanes \& Bicycles}

Although previous studies in single-category skeleton detection have mainly focused on pedestrians, animals, and cars, lanes and bicycles, two key categories in autonomous driving, have received less attention. In this work, we reformulate lane detection as a skeleton detection problem. Additionally, due to the lack of publicly available datasets with annotated bicycle keypoints, we introduce keypoint annotations for bicycle skeleton detection in the MS COCO dataset. These tasks are detailed in this section and later integrated into our multi-task skeleton detection framework.

\subsection{Skeleton-Based Lane Detection}
\label{sec:gt}


\begin{figure*}
    \centering
   
    \includegraphics[width=0.95\textwidth, clip]{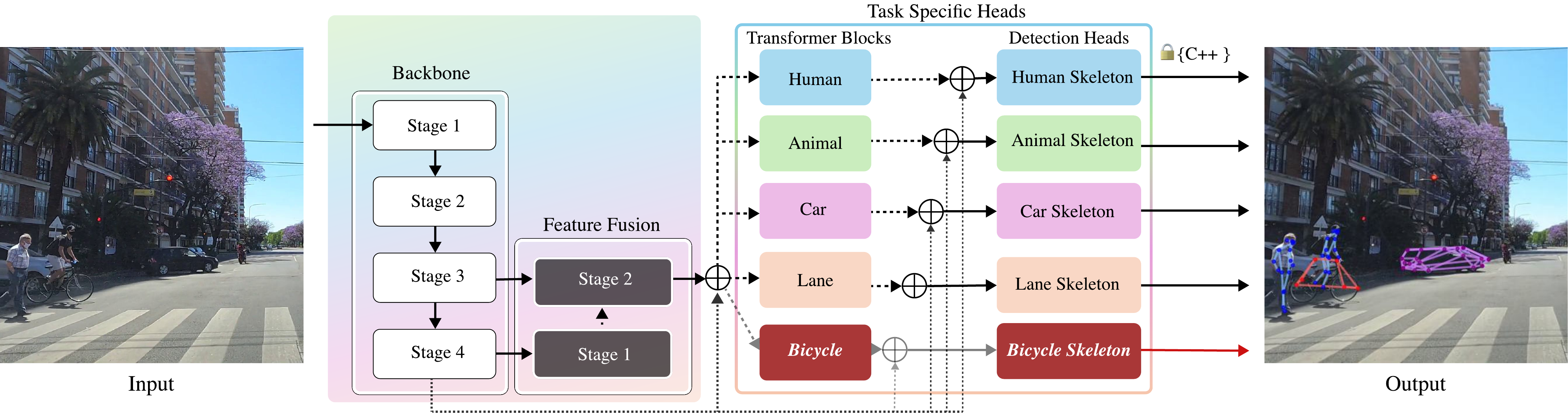}

    \caption{
    Overview of our framework: Our network is designed to detect skeletons for pedestrians, animals, cars, bicycles, and lanes. A feature fusion stage is incorporated after the backbone and task-specific transformer blocks to enhance overall performance. In our schematic diagrams, dashed and dotted paths represent different architectural variations (e.g., configurations without transformer or FPN). The $\oplus$ nodes indicate the merging of two inputs; when only one input is connected to the output, this notation is used solely for visualization purposes to illustrate different variants of our architecture and should not be interpreted as a summation or other arithmetic operation.}
    \label{fig:overview}
\end{figure*}



Our framework generalizes skeleton detection beyond tasks with explicitly defined keypoints. Rather than treating lane detection as a separate problem, we reframe it within our skeleton-based approach by representing lane lines as fixed keypoints. This enables seamless integration of lane detection alongside other object classes, even when keypoints lack direct semantic meaning. This unification highlights the adaptability of our method in handling diverse detection tasks within a single, cohesive framework.


Given an input RGB image $\mathbf{I} \in{\mathbb{R}}^{3 \times H \times W}$, the objective of our model is to produce a set of lanes $L=\{l_1, l_2, ..., l_N\}$, where N is the number of lanes detected in the image. Each lane $l_i$ is expressed as a fixed number of ordered 2D-pixel coordinates, that is, keypoints: $l_i = [(x_i^1, y_i^1), (x_i^2, y_i^2), ..., (x_i^M, y_i^M)]$, where $x$ and $y$ are coordinates, M is the predetermined cardinality of the set of output points. Unlike other tasks in the framework, lane keypoints lack a clear semantic meaning. For example, while human keypoints have specific semantic meanings (e.g., eyes, nose), lane keypoints represent structural points that are harder to define semantically.

To ensure compatibility with our designed lane keypoints for differently formatted lane datasets, and address the ambiguity of the concept of lane keypoints, we carefully processed the ground truth annotations during the training phase.  
Current lane data sets have different forms of lane-point annotations. The OpenLane data set~\cite{chen2022persformer} densely annotates 2D lane points. This annotation format ends up in a random number of lane point annotations, which is incompatible with a uniform keypoints representation for all lane instances. We kept the starting and ending points and interpolated the rest of the annotated points for each lane into a fixed number of uniformly distributed points in pixel distance. Beyond framework compatibility assurance, such spatial geometrical constraints imposed on the adjacent keypoints also mitigate the inherent lane keypoint ambiguity compared to semantic keypoints such as human joints due to the lack of natural visual cues. Besides, such equal pixel space representation eliminates any fixed vertical intervals, which favors near-flat lane annotation and detection. 

\subsection{Bicycle Skeleton Annotation}
We used a subset of the MS COCO dataset, which includes a wide variety of objects, such as bicycles. From this, we selected 1,557 images featuring bicycles and manually annotated them for bicycle keypoint detection. Each bicycle was annotated with six keypoints: the back and center of the rear wheel, the front and center of the front wheel, the seat, and the center of the handlebars, resulting in 2,439 annotated instances. Of these, 1,245 images were randomly assigned to the training set, while the remaining 312 images were reserved for testing.

\section{Multi-Category Skeleton Detection in a Unified Framework}
\label{sec:unified_framework}



Drawing inspiration from OpenPifPaf \cite{kreiss2021openpifpaf}, we propose a bottom-up network architecture that employs Composite Intensity Fields (CIF) for keypoint detection and Composite Association Fields (CAF) for part association. Although OpenPifPaf has been validated on single-category tasks (e.g., pedestrians, animals, and cars), its design presents limitations when directly applied to multi-category skeleton detection. Similar constraints have been observed in other single-category \cite{ye2022superanimal, xu2022vitpose} and category-agnostic \cite{shi2023matching, hirschorn2023pose, xu2022pose, nguyen2024escape, yang2025x} skeleton detection networks.

\paragraph{Backbone.}
OpenPifPaf employs ShuffleNet and ResNet as its backbone networks. However, these architectures incorporate Batch Normalization, which prior research \cite{chang2019domain, xie2019intriguing, schneider2020improving} has shown to capture domain-specific characteristics. In our work, the absence of a unified dataset for the categories of interest necessitated the use of four distinct datasets, potentially exacerbating domain bias issues. Previous work \cite{liu2022convnet} demonstrated that substituting Batch Normalization with Layer Normalization in ResNet can lead to performance degradation, whereas ConvNeXt remains robust under this modification. We extend this investigation to ShuffleNet and report two key findings. First, despite similarities between the domains of datasets (e.g., MS COCO and Animal Pose), using Batch Normalization can result in a significant performance drop in a multi-task setting. Second, replacing Batch Normalization with Layer Normalization needs further investigation. Consequently, for our experiments, we started with ShuffleNet, and then opted for backbones such as Swin, ConvNeXtv2, and ClipConvNeXt which were originally designed without Batch Normalization layers.

\paragraph{Handling Diverse Object Scales.}
Many skeleton detection methods \cite{kreiss2021openpifpaf, ye2022superanimal, xu2022vitpose} were originally developed for single-category applications and therefore lack specialized modules to address objects at varying scales. This challenge is also evident in category-agnostic networks \cite{shi2023matching, hirschorn2023pose, xu2022pose, nguyen2024escape, yang2025x}. While scale variation poses a challenge even in single-category settings, it becomes more pronounced when extending to multiple categories, as illustrated by the diverse scale distributions summarized in Table \ref{table:datasets}. To address this issue, we employ two strategies. First, recognizing that object detectors often integrate Feature Pyramid Networks (FPN) \cite{lin2017feature} to capture and fuse multi-scale features, we augment our backbone features with an FPN module. Second, we utilize mosaic augmentation \cite{bochkovskiy2020yolov4} with a sampling strategy designed to increase the likelihood of incorporating images containing larger-scale objects.

\paragraph{Lack of Task-Specific Layers.}
Applying CIF and CAF directly on top of a backbone network is often sufficient in single-category skeleton detection. 
However, when dealing with multiple object categories, a generic approach becomes suboptimal due to the significant differences in the appearance, structure, and keypoint requirements across the classes. To address this, we incorporate a transformer block before each network head to refine feature extraction for each specific category. We adopt the Swin Transformer block for two primary reasons. First, the backbone’s large receptive field ensures that local windowed attention (using a 7×7 window) can effectively capture relevant details while mitigating the quadratic computational cost associated with global attention. Second, the extensive receptive field of the backbone supports the aggregation of global contextual information despite the localized attention mechanism.

\color{black}

\section{Experiments}

\begin{table}[t!]
\renewcommand{\arraystretch}{1.3} 
\caption{Summary of Datasets Used in Experiments: The table presents an overview of the datasets, with the scale metric defined as the average bounding box area of each sample normalized by the corresponding image area.}
\label{table:datasets}
\centering
\begin{adjustbox}{max width=\linewidth}
\begin{tabular}{cccccc}
    \toprule
    \textbf{Dataset} & \textbf{Year} & \textbf{Images} & \textbf{Instances} & \textbf{Scale (\%)} & \textbf{Category} \\ 
    \midrule
    MS COCO \cite{lin2014microsoft}          & 2014          & 200K            & 150K                 & 8.10 & Human             \\ 
    AnimalPose \cite{cao2019cross}       & 2019          & 4K              & 6K                   & 37.55 & Animal            \\ 
    AwA-Pose \cite{banik2021novel}       & 2021          & 10K             & 10K                  & 52.31 & Animal            \\ 
    ApolloCar3D \cite{song2019apollocar3d} & 2018          & 5K              & 60K                  & 0.35 & Car               \\ 
    OpenLane \cite{chen2022persformer}    & 2022          & 200K            & 660K                   & 6.03 & Lane              \\ 
    {MS COCO Bicycle (Ours)} & 2025 & 1.5K & 2.4K & 4.24 & Bicycle \\
    \bottomrule
\end{tabular}
\end{adjustbox}
\end{table}

\subsection{Datasets}
In this paper, we use multiple datasets to evaluate our approach across different object categories, including humans, animals, cars, bicycles, and lanes. This setup ensures a comprehensive assessment of our model’s ability to handle diverse skeleton structures within a unified framework. For human skeleton detection, we use the MS COCO keypoint detection set \cite{lin2014microsoft}, which provides annotations for 17 body parts across 80,000 instances. For animals, we use the AnimalPose dataset \cite{cao2019cross}, which includes keypoint annotations for five species, and the AwA-Pose dataset \cite{banik2021novel}, which covers 35 quadrupedal species with corresponding visual and semantic attributes.
For vehicle keypoint detection, we rely on the ApolloCar3D dataset \cite{song2019apollocar3d}, which contains 60,000 annotated samples of cars in urban traffic conditions. Lane detection is evaluated on the OpenLane dataset \cite{chen2022persformer}, which presents various challenging driving scenarios.

\textbf{MS COCO keypoint detection set \cite{lin2014microsoft}} 
The MS COCO dataset is a widely used benchmark for human pose estimation. It contains 51,000 images with over 80,000 annotated instances of people. The dataset includes keypoint annotations for 17 body parts, providing a rich resource for training and evaluating pose estimation models in various conditions such as occlusions, different viewpoints, and diverse backgrounds.

\textbf{AnimalPose \cite{cao2019cross}} 
The dataset comprises 4,000 images featuring 6,000 instances of animals in various poses. It offers comprehensive keypoint annotations for five animal species, including dogs, cats, horses, sheep, and cows. The images in the dataset are sourced from a variety of platforms, including the Internet and other existing datasets.

\textbf{Animals with Attributes (AwA-Pose) \cite{banik2021novel}} 
This dataset comprises 10,000 images representing 50 animal species. From this collection, 35 quadrupedal species were selected, each accompanied by corresponding visual and semantic attributes. Only images featuring a single animal in each instance have been included, ensuring a focused approach for pose estimation tasks.

\textbf{ApolloCar3D Dataset \cite{song2019apollocar3d}} 
The ApolloCar3D dataset comprises 5,000 images with 60,000 annotated samples. It focuses on vehicle keypoint extraction, offering high-resolution images from urban traffic scenes captured under diverse conditions. The dataset is specifically designed to evaluate car part localization and other tasks related to autonomous driving and vehicle recognition.

\textbf{OpenLane \cite{chen2022persformer}} 
The OpenLane dataset includes 160,000 images for training and 40,000 images for validation. It is designed for lane detection in autonomous driving scenarios and features a variety of challenging situations, including curves, intersections, night driving, extreme weather conditions, and lane merges and splits.

\textbf{MS COCO Bicycle keypoint detection set \cite{lin2014microsoft}} 
We used a subset of the MS COCO dataset, which includes a wide variety of objects, such as bicycles. From this, we selected 1,557 images featuring bicycles and manually annotated them for bicycle keypoint detection. Each bicycle was annotated with six keypoints: the back and center of the rear tire, the front and center of the front tire, the seat, and the center of the handlebars, resulting in 2,439 annotated instances. Of these, 1,245 images were randomly assigned to the training set, while the remaining 312 images were reserved for testing.

\subsection{Experiments on Lane Detection}
This section shows the results of our proposed skeleton-based keypoint detection method (explained in \Cref{sec:gt}).
\textbf{Experiments on OpenLane Dataset \cite{chen2022persformer}. } For the OpenLane dataset, which does not provide an open test set, we evaluate our method on the OpenLane validation set, following the approach in \cite{chen2023generating}, as shown in \Cref{tab:Openlane}. Our small model, using the ShuffleNetV2K16 backbone, outperforms all alternatives prior to CondLSTR \cite{chen2023generating}, as well as the small version of CondLSTR (using a ResNet-18\cite{he2016deep} backbone), achieving an F1 score of 60.6\%. Our medium-sized model, using ShuffleNetV2K32, achieves state-of-the-art performance with an F1 score of 63.6\%, outperforming the similarly-sized CondLSTR-M \cite{chen2023generating} (ResNet-34 backbone) by 1.2 points, and even the larger CondLSTR-L \cite{chen2023generating} (ResNet-101 backbone) by 0.2\%. Moreover, in order to align with multi-task experiments in \cref{sec:multi_task_sec}, we include additional experiments using recent backbones: ConvNeXt and Swin-L with Feature Pyramid Networks (FPN) \cite{lin2017feature}. Both of the two models surpass all other alternatives by significant margins 4.1 and 8.1 points over CondLSTR-L. This highlights the efficiency of our model and the effectiveness of the fixed-number keypoints lane representation.

Following \cite{chen2022persformer,pan2018spatial}, we adopt $F_1$ measure for evaluation. Each lane instance is treated as a line with a fixed width of 30 pixels, and intersection-over-union (IoU) with a threshold of 0.3 between the prediction and the matched ground truth is applied to judge whether the prediction is a True Positive (TP) or False Positive (FP). The $ F_1$ is calculated as:
\begin{equation}
Precision = \frac{N_{TP}}{N_{TP}+N_{FP}}, \quad Recall = \frac{N_{TP}}{N_{TP}+N_{FN}}, \quad  F_1 = \frac{2 \times Precision \times Recall}{Precision + Recall}.
\end{equation} 
We report $F_1$ measure for both overall scenes and split cases for both datasets.

Additionally, our model achieves state-of-the-art performance in challenging conditions, including extreme weather and nighttime scenarios, with F1 scores of 58.3\% and 61.0\% for our middle-sized model\color{black}, respectively. This surpasses the previous state-of-the-art CondLSTR-L \cite{chen2023generating} by 3.2 and 3.7 points. Our model with Swin-L + FPN backbone consistently achieves the highest scores on all challenging scenarios up to 11.9 points over CondLSTR-L. This underscores the robustness of our approach, especially in variable lighting conditions and under partial occlusion or blurring from raindrops on the windshield or water reflections on the road surface.

\begin{table*}[tb]
  \caption{OpenLane results, the sizes of CondLSTR \cite{chen2023generating} are estimated based on \cite{liu2021condlanenet}, which has similar architecture and the corresponding backbones used. The best results are \textbf{bolded}, and the second-best results are \underline{underlined}.}
  \label{tab:Openlane}
  \centering
  \begin{adjustbox}{max width=\textwidth}
  \begin{tabular}{@{}lcccccccc@{}}
    \toprule
    \textbf{Method} & \textbf{All} & \textbf{Up\&Down} & \textbf{Curve} & \textbf{Extreme Weather} & \textbf{Night} & \textbf{Intersection} & \textbf{Merge\&Split} & \textbf{Parameters (M)} \\
    \midrule
    LaneATT-S \cite{9577584}  & 28.3  & 25.3 &25.8&32.0&27.6&14.0&24.3 & 13.3\\
    LaneATT-M \cite{9577584} & 31.0 &28.3 &27.4&34.7&30.2&17.0&26.5 & 23.4\\
    PersFormer \cite{chen2022persformer} & 42.0 & 40.7&46.3&43.7&36.1&28.9&41.2 &54.9\\
    CondLaneNet-S \cite{liu2021condlanenet} & 52.3 &55.3 &57.5&45.8&46.6&48.4&45.5&12.1\\
    CondLaneNet-M \cite{liu2021condlanenet} &55.0&58.5 &59.4&49.2&48.6&50.7&47.8&22.2\\
    CondLaneNet-L \cite{liu2021condlanenet} &59.1&62.1 &62.9&54.7&51.0&55.7&52.3&49.9\\
    CondLSTR-S \cite{chen2023generating} &60.1&56.2 &63.9&51.5&51.0&54.5&62.5& $\approx{15.0}$\\
    CondLSTR-M \cite{chen2023generating} &62.0 &59.1 &65.4 &53.4&54.1&57.3&63.2&$\approx{25.0}$\\
    CondLSTR-L \cite{chen2023generating} &63.4 & \underline{62.2} & 67.0 & 55.1 & 57.3 & 58.5 & 65.8 &$\approx{50.0}$\\
    \hline
    OpenPifPaf (ShuffleNetv2K16) & 60.6 & 56.9 & 62.1 & 54.3 & 59.6 & 53.6 & 59.5 &\textbf{10.5}\\
    OpenPifPaf (ShuffleNetv2K32) & 63.6  & 60.5 & 65.2& 58.3 & 61.0 & 54.7 & 64.0 & 30.5\\
    OpenPifPaf (ConvNeXt) & \underline{67.5} & 61.8 & \underline{73.8} & \underline{59.6} & \underline{64.7} & \underline{59.0} & \underline{66.5} & 88.5 \\
    OpenPifPaf (Swin-L + FPN) & \textbf{71.5} & \textbf{66.8} & \textbf{77.7} & \textbf{63.9} & \textbf{69.2} & \textbf{61.2} & \textbf{70.7} & $\approx{220}$ \\
  \bottomrule
  \end{tabular}
  \end{adjustbox}
\end{table*}

\textbf{Experiments on CULane Dataset \cite{pan2018spatial}. }
Our sparse point representation based method requires precise regression of the lane keypoints, which relies on the training dataset scale and annotation quality to ensure its generalization ability. Although CULane contains only half as many images as OpenLane, and 15\% of the frames do not contain any lane annotations, our method still achieves competitive results with the state-of-the-art model CLRerNet-DLA-34 \cite{honda2024clrernet}, achieving a $F_1$ measure of 80.01\%. And, our method also outperforms the baseline keypoint-based model GANet \cite{wang2022keypoint} which also requires precise location regression of lane points for both the overall set and most of the challenging scenarios including Crowded, Shadow, No line, Arrow, and Night scenes. 

\begin{figure*}[t!]
  \centering
  \includegraphics[height=8cm]{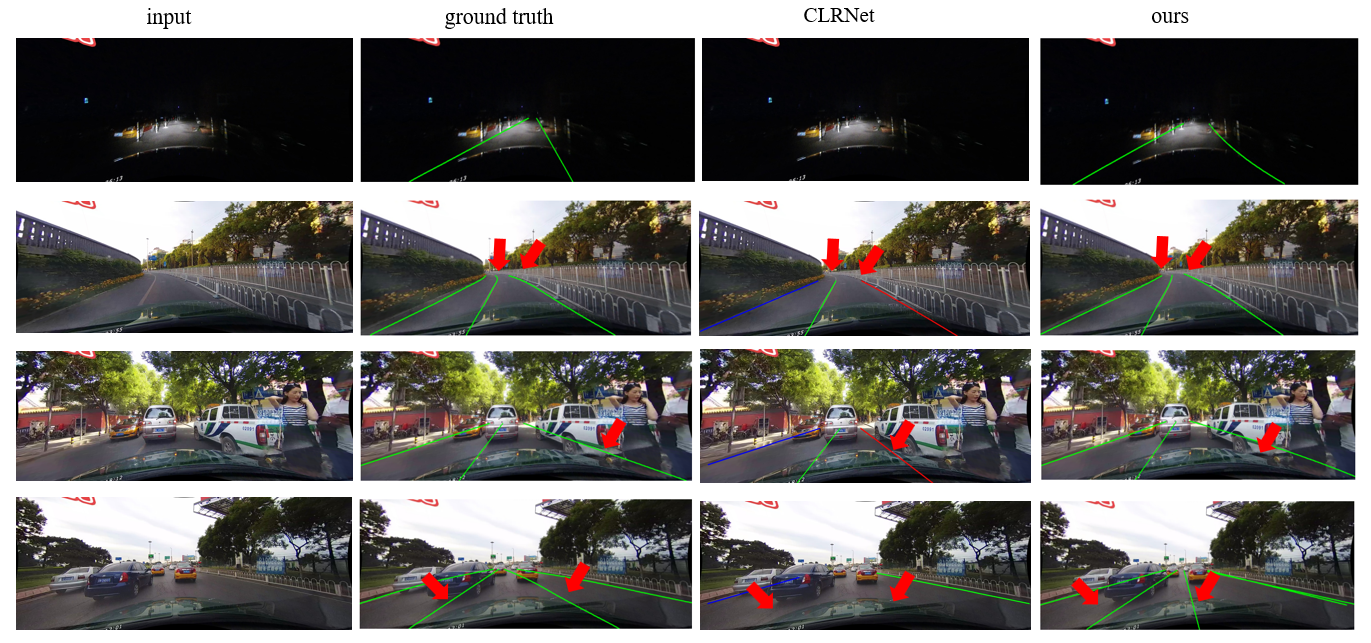}
  \caption{Qualitative results of our method ($4^{th}$ column) on CULane test set, compared with CLRNet \cite{zheng2022clrnet} ($3^{rd}$ column). The first row shows an example of a night scene, the second row contains curved lane lines, and the third and fourth rows showcase crowded scenes with severe occlusion. The red arrows point at the parts where our method clearly generates better predictions.
  }
  \label{fig:vis}
\end{figure*}

On top of the overall performance, our model also achieves the best performances for crowded scenes with partial occlusions in keypoint-based methods group. This demonstrates the effectiveness of our intuition of addressing the partial occluded lanes challenge for keypoint-based methods by using composite fields to jointly predict keypoint locations and associations. Some of the qualitative results on challenging scenarios of CULane test set are also visualized and compared with the current $3^{rd}$ ranking method, CLRNet \cite{zheng2022clrnet}, which has the pre-trained models available, as shown in \Cref{fig:vis}. The better outcomes on these challenging scenes further show the effectiveness of our method.
\begin{table*}[t!]
  \caption{Quantitative results on CULane in $F_1$ score (\%). The best performances are marked in \textbf{bold}, and the local best performances for keypoint based methods are marked in \underline{underline}. 
  }
  \label{tab:culane}
  \centering
  \begin{adjustbox}{max width=\textwidth}
  \begin{tabular}{@{}lcccccccccc@{}}
    \toprule
    Method &  Total & Normal & Crowded & Dazzle & Shadow & No line & Arrow &Curve & Cross &Night \\
    \midrule
    LaneATT-S\cite{9577584}  & 75.13 & 91.17 &72.71&65.82&68.03&49.13&87.82&63.75&1020&68.58\\
    LaneATT-M\cite{9577584} & 76.68&92.14 &75.03&66.47&78.15&49.39&88.38&67.72&1330&70.72\\
    LaneATT-L\cite{9577584} & 77.72 & 91.74&76.16&69.47&76.31&50.46&86.29&64.05&1264&70.81\\
    CondLaneNet-S\cite{liu2021condlanenet}&78.14 &92.87 &75.79 &70.72&80.01&52.39&89.37&72.40&1364&73.23\\
    CondLaneNet-M\cite{liu2021condlanenet}&78.74 &93.38&77.14&71.17&79.93&51.85&89.89&73.88&1387&73.92\\
    CondLaneNet-L\cite{liu2021condlanenet} &79.48& 93.47&77.44&70.93&80.91&54.13&90.16&75.21&1201&74.80\\
    CLRNet-DLA34\cite{zheng2022clrnet}&80.47&93.73&79.59&\textbf{75.30}&82.51&54.58&90.62&74.13&1155&75.37\\
    CondLSTR-L\cite{chen2023generating}&80.77&94.17&79.90&75.43&80.99&55.00&90.97&76.87&1047&75.11\\
    CLRerNet-DLA34\cite{honda2024clrernet} & \textbf{81.43}&\textbf{94.36} &\textbf{80.62} &75.23 &84.35 &57.31 &91.17&79.11&1540&\textbf{76.92}\\
    \hline
    \hline
    \textbf{Keypoint-based}\\
    FoloLane\cite{qu2021focus}&78.80&92.70&77.80&\underline{75.20}&79.30&52.10&89.00&69.40&1569&74.50\\
    GANet-S\cite{wang2022keypoint}&78.79&93.24&77.16&71.24&77.88&53.59&89.62&75.92&1240&72.75\\
    GANet-M\cite{wang2022keypoint}&79.39&93.73&77.92&71.64&79.49&52.63&90.37&76.32&1368&73.67\\
    GANet-L\cite{wang2022keypoint}&79.63&93.67&78.66&71.82&78.32&53.38&89.86&77.37&1352&73.85\\
    RCLane-M\cite{xu2022rclane}&80.03&93.59&78.77&72.44&\underline{\textbf{84.37}}&52.77&90.31&78.39&\underline{\textbf{907}}&73.96\\
    RCLane-L\cite{xu2022rclane}& \underline{80.50}&\underline{94.01}&79.13&72.92&81.16&53.94&90.51&\underline{\textbf{79.66}}&931&75.10\\
    OpenPifPaf (ShuffleNetv2K16) &78.50 &93.16 &78.61&73.77&80.24&57.23&90.61&77.52&1342&75.19\\
    OpenPifPaf (ShuffleNetv2K32) &80.01 & 93.92&\underline{79.79}&71.77&81.67&\underline{\textbf{58.28}}&\underline{\textbf{91.66}}&76.43&1210&\underline{76.54}\\
  \bottomrule
  \end{tabular}
  \end{adjustbox}
\end{table*}

\begin{table*}[h]
  \caption{Detailed quantitative results for multi-category networks on the OpenLane dataset categorized by different extreme conditions.}
  \label{tab:Openlane_2}
  \centering
  \begin{adjustbox}{max width=\textwidth}
  \begin{tabular}{@{}lcccccccccc@{}}
    \toprule
    \textbf{Backbone} & \textbf{\begin{tabular}{@{}c@{}}
        Mosaic\\Aug.
        \end{tabular}} & \textbf{FPN} & \textbf{\begin{tabular}{@{}c@{}}
        Attention\\Head 
        \end{tabular}}  & \textbf{All} & \textbf{Up\&Down} & \textbf{Curve} & \textbf{Extreme Weather} & \textbf{Night} & \textbf{Intersection} & \textbf{Merge\&Split}  \\
    \midrule
    \multirow{4}{*}{ConvNeXtv2} & \xmark & \xmark & \xmark & 55.2 & 47.4 & 60.5 & 48.5 & 55.3 & 46.3 & 54.9 \\
     & \cmark & \cmark & \xmark & 63.9 & 57.9 & 69.2 & 59.5 & 65.6 & 52.2 & 63.6 \\
     & \cmark & \xmark & \cmark & 61.5 & 56.9 & 67.7 & 56.4 & 60.8 & 50.3 & 59.8 \\
     & \cmark & \cmark & \cmark & 67.7 & 62.7 & 70.2 & 61.0 & 67.9 & 59.2 & 67.3 \\

     \bottomrule

     \multirow{4}{*}{ClipConvNeXt} & \xmark & \xmark & \xmark & 67.0 & 64.5 & 72.4 & 60.3 & 65.3 & 58.4 & 64.0 \\
     & \cmark & \cmark & \xmark & 67.6 & 62.6 & 72.3 & 58.7 & 66.4 & 58.5 & 67.1 \\
     & \cmark & \xmark & \cmark & 65.7 & 63.0 & 71.9 & 56.8 & 62.7 & 57.3 & 62.6 \\
     & \cmark & \cmark & \cmark & 67.7 & 64.4 & 70.9 & & 66.9 & 60.5 & 68.3 \\

     \bottomrule
     
     \multirow{4}{*}{SwinL} & \xmark & \xmark & \xmark & 69.1 & 67.8 & 76.7 & 62.8 & 68.7 & 59.3 & 65.9 \\
     & \cmark & \cmark & \xmark & 65.4 & 61.4 & 72.4 & 60.1 & 61.7 & 53.0 & 61.0 \\
     & \cmark & \xmark & \cmark & 71.0 & 68.0 & 77.0 & 64.6 & 69.6 & 61.4 & 67.9 \\
     & \cmark & \cmark & \cmark & 69.4 & 64.4 & 74.3 & 62.9 & 67.9 & 60.7 & 69.0 \\

  \bottomrule
  \end{tabular}
  \end{adjustbox}
\end{table*}

\textbf{Sample Selection}.
To evaluate the impact of different downsampling strategies for processing ground truth, we conducted an ablation study on three methods using ShuffleNetV2K16 on the OpenLane dataset, targeting 24 keypoints per lane:
\begin{itemize}
    \item \textbf{Random Downsampling ($A$):} Selects 24 points randomly from the lane line.
    \item \textbf{Fixed Vertical Distance Downsampling ($B$):} Applies a 20-pixel interval for point selection. For lanes with more than 24 points, random downsampling is applied to reduce the count; for lanes with fewer than 24 points, interpolation is used to reach the target.
    \item \textbf{Even Downsampling with Fixed Endpoints ($C$):} Retains the start and end points of the lane, evenly distributing the remaining keypoints, as described in \Cref{sec:gt}.
\end{itemize}

The quantitative results on the CULane test set are presented in \Cref{tab:ablation}. Random downsampling ($A$) performed poorly, achieving only a 73.24 $F_1$ score due to poor keypoint discrimination. Fixed vertical intervals ($B$) improved keypoint learnability by introducing consistency but was constrained by adjustments to maintain the keypoint count. Method ($C$), which evenly distributes points while preserving endpoints, achieved the best results by reducing keypoint ambiguity and maintaining lane structure, confirming its superiority for this task.

\begin{figure*}[tb]
  \centering
  \includegraphics[height=14.5cm]{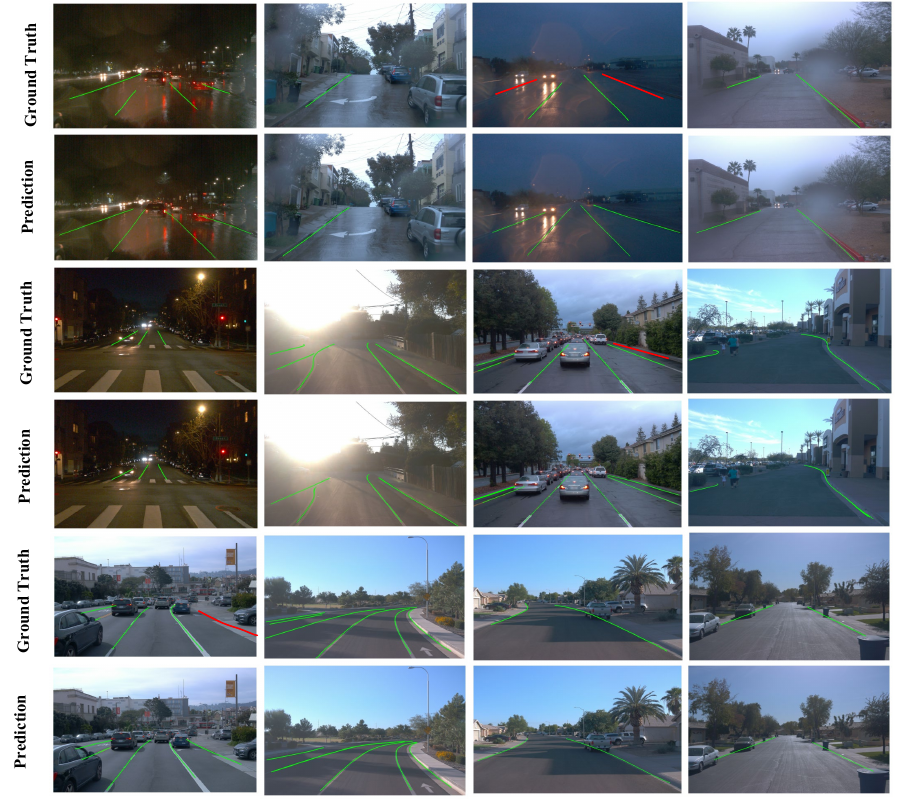}
  \caption{Qualitative results of our method on OpenLane validation set, including crowded scenes with occlusion, extreme weather, dazzling sunshine, converging lanes, night-time driving scenarios, and curve-shaped lanes. Missing annotations in the ground truth are marked as red lines in the images.
  }
  \label{fig:vis-openlane}
\end{figure*}

\begin{table}[h!]
  \centering
  \caption{Ablation study on downsampling method}
  \label{tab:ablation}
  \renewcommand{\arraystretch}{1.2} 
  \small 
  \begin{tabular}{@{}l c@{}}
    \toprule
    \textbf{Sample Selection} & $F_1$ \\ 
    \midrule
    Random Downsampling & 73.24 \\
    Fixed Vertical Distance Downsampling & 76.93 $\uparrow^{3.69}$ \\
    Even Downsampling with Fixed Endpoints & 78.50 $\uparrow^{1.57}$ \\
    \bottomrule
  \end{tabular}
\end{table}

\subsection{Experiments on Multi-Tasks}
\label{sec:multi_task_sec}

In this section, we integrate four detection tasks (e.g. pedestrian, animal, car, and lane detection) into a unified multi-task framework. Bicycle detection is omitted to assess the framework’s generalization and adaptability to new categories. We investigate the impact of various mitigation strategies and present experimental comparisons that demonstrate robust performance across all four tasks.

\begin{table}[t!]
\renewcommand{\arraystretch}{1.1} 
\caption{Performance of ShuffleNet Variants on Pedestrians and Animals Keypoint Detection. Metrics include Average Precision (AP) and the difference ($\Delta$) for single (S) and multi-task (M) setups.}
\label{table:shufflenet_comparison}
\footnotesize
\begin{adjustbox}{max width=\textwidth}
\centering
\begin{tabular}{cccccc}  
    \toprule
    \begin{tabular}{@{}c@{}}
    \textbf{Normalization}
    \end{tabular} & \begin{tabular}{@{}c@{}}
    \textbf{Pedestrian}
    \end{tabular} & $\Delta$ & 
    \begin{tabular}{@{}c@{}}
    \textbf{Animal}
    \end{tabular} & \textbf{$\Delta$} & \textbf{S/M} \\ 
    \midrule

    \multirow{3}{*}{\begin{tabular}{@{}c@{}}
    Batch\\Normalization
    \end{tabular}} & 71.8 & -- & -- & -- & S \\
    & -- & -- & 49.5 & -- & S \\ 
    & 63.4 & -8.4 & 38.6 & -10.9 & M \\ \midrule

    \multirow{3}{*}{\begin{tabular}{@{}c@{}}
    Layer\\Normalization
    \end{tabular}} & 70.2 & -- & -- & -- & S \\
    & -- & -- & 49.2 & -- & S \\
    & 65.9 & -4.3 & 47.8 & -1.4 & M \\ 

    \bottomrule
\end{tabular}
\end{adjustbox}
\end{table}

\begin{table*}[!t]
    \renewcommand{\arraystretch}{1.3} 
    \caption{Performance comparison of backbone architectures in single (S) and multi-task (M) settings. This table summarizes the effects of Mosaic augmentation,  FPN, and transformer block on model performance, measured by Average Precision (AP) for pedestrian, animal, and car detection, F1 score for lane detection, and overall mean score across all tasks in a multi-category skeleton detection setting.  \cmark indicates the feature is applied, while \xmark means it is not. Experiments marked with * indicate the use of FPN.}
    \label{table:backbone_comparison}
    \small
    \centering
    \begin{adjustbox}{max width=\textwidth}
    \begin{tabular}{lccccccccccc} 
        \toprule
        \textbf{Backbone} & \textbf{\begin{tabular}{@{}c@{}}
        S/M
        \end{tabular}} & \textbf{\begin{tabular}{@{}c@{}}
        Mosaic\\Aug.
        \end{tabular}} & \textbf{FPN} &
        \textbf{\begin{tabular}{@{}c@{}}
        Transformer\\Block
        \end{tabular}} & \textbf{\begin{tabular}{@{}c@{}}
        Params\\(M)
        \end{tabular}} & \textbf{\begin{tabular}{@{}c@{}}
        Pedestrian\\(AP\%)
        \end{tabular}} & \textbf{\begin{tabular}{@{}c@{}}
        Animal\\(AP\%)
        \end{tabular}} & \textbf{\begin{tabular}{@{}c@{}}
        Car\\(AP\%)
        \end{tabular}} & \textbf{\begin{tabular}{@{}c@{}}
        Lane\\(F1\%)
        \end{tabular}} & \textbf{\begin{tabular}{@{}c@{}}
        Mean\\Score
        \end{tabular}} \\
        \midrule
        \multirow{1}{*}{SuffleNetK30} & M & \xmark & \xmark & \xmark & 30 & 45.6 & 33.7 & 56.7 & 33.4 & 42.8  \\
        \midrule
        \multirow{5}{*}{ConvNeXtv2} & S & \xmark & \xmark & \xmark & 88 & \textbf{70.5} & 44.4 & 69.8 & 67.5 & 63.0  \\ 
        & M & \xmark & \xmark & \xmark & 88 & 63.4 & 47.9 & 70.8 & 55.2 & 59.3  \\
        & M & \cmark & \cmark & \xmark & 99 & 67.0 & 53.6 & 71.9 & 63.9 & 64.1  \\ 
        & M & \cmark & \xmark & \cmark & 188 & 66.3 & 55.6 & 70.2 & 61.5 & 63.4  \\ 
        & M & \cmark & \cmark & \cmark & 199 & 69.0 & \textbf{55.9} & \textbf{72.9} & \textbf{67.7} & \textbf{66.4}  \\ 
        \midrule
        \multirow{5}{*}{ClipConvNeXt} & S & \xmark & \xmark & \xmark & 88 & 69.3 & 51.6 & 70.0 & 64.6 & 63.9  \\ 
        & M & \xmark & \xmark & \xmark & 88 & 64.8 & 49.5 & 70.6 & 67.0 & 63.0  \\
        & M & \cmark & \cmark & \xmark & 99 & 66.9 & 52.2 & 71.6 & 67.6 & 64.6  \\ 
        & M & \cmark & \xmark & \cmark & 188 & 66.5 & 53.3 & 70.6 & 65.7 & 64.0  \\ 
        & M & \cmark & \cmark & \cmark & 199 & \textbf{68.1} & \textbf{53.5} & \textbf{72.7} & \textbf{67.7} & \textbf{65.5}  \\
        \midrule
        \multirow{5}{*}{Swin L} & S & \xmark & \xmark & \xmark & 196 & \textbf{70.0} & 51.5 & 71.8 & \textbf{71.5*} & \textbf{66.2}  \\ 
        & M & \xmark & \xmark & \xmark & 196 & 64.7 & 41.0 & 69.7 & 69.2 & 61.2  \\
        & M & \cmark & \cmark & \xmark & 221 & 69.8 & 41.0 & 70.8 & 64.4 & 61.5  \\ 
        & M & \cmark & \xmark & \cmark & 424 & 67.8 & 50.4 & 70.7 & 71.0 & 65.0  \\ 
        & M & \cmark & \cmark & \cmark & 445 & 66.1 & \textbf{53.6} & \textbf{72.5} & 69.4 & 65.4  \\ 
        \bottomrule
    \end{tabular}
    \end{adjustbox}
\end{table*}

        


      

\textbf{Backbone Effect.} As an initial step, we began with the baseline OpenPifPaf \cite{kreiss2019pifpaf, kreiss2021openpifpaf} framework, utilizing the ShuffleNetv2 \cite{ioffe2015batch} backbone. The network was first trained from scratch on the COCO human keypoint dataset, and the resulting weights were then used to train on the AnimalPose dataset for detecting animal keypoints. For the multi-task setup, the network was initialized with the weights obtained from human keypoint training. Results from this setup are presented in \Cref{table:shufflenet_comparison}. However, the original ShuffleNetv2 architecture proved insufficient for multi-category keypoint extraction, with an average performance drop of 9.7 points across tasks compared to single-task training.
Our studies revealed that a key limitation was ShuffleNetv2's reliance on batch normalization, which normalizes feature maps across each batch. In a multi-task setup, this can hinder performance as it learns data distributions within a mini-batch, making adaptation to tasks with different distributions challenging (see \Cref{sec:unified_framework}).

To address this, batch normalization was replaced with layer normalization, which normalizes feature maps across the feature dimension rather than the batch. As shown in \Cref{table:shufflenet_comparison}, this adjustment significantly improved multi-category keypoint detection. For example, while the original ShuffleNetv2 with batch normalization showed a performance drop of over 8 points per task, replacing it with layer normalization reduced the drop to 4.3 points for humans and 1.4 for animals.


Modifying the normalization method in the ShuffleNet backbone improved multi-task performance but resulted in significant drops in single-task keypoint detection. For instance, using instance normalization led to a 2 and 6 points decrease in AP scores for human and animal detection, respectively. Layer normalization similarly caused a 1 point average performance decline. The performance degradation was more pronounced when ShuffleNet was applied to all four tasks, as shown in Table \ref{table:backbone_comparison}. Specifically, when ShuffleNet was trained with layer normalization across the four tasks, the results indicated that even with normalization modifications, ShuffleNet is not well-suited for multi-task skeleton detection. Building on this observation, we explored recent backbones with built-in layer normalization, such as ConvNeXt and Swin Transformer models. These models, pre-trained on large-scale datasets, are better suited for our study. Additionally, we evaluated ConvNeXt pre-trained on the LAION dataset \cite{schuhmann2021laion}, referred to as ClipConvNeXt, to assess the effectiveness of models optimized for layer normalization.

For ConvNeXtv2, ClipConvNeXt, and Swin-L, the first row in Table \ref{table:backbone_comparison} shows their performance on a single task ("S"), while the second row presents their baseline multi-task ("M") performance without additional modules. As seen in \Cref{table:backbone_comparison}, their multi-task results outperform ShuffleNetK30 by 16.5, 20.2, and 18.38 points, respectively. This substantial improvement motivated us to focus further experiments on these backbones.

\begin{figure*}[t!]
    \centering

    \begin{subfigure}[b]{0.45\textwidth}
        \centering
        \includegraphics[height=5cm, width=\linewidth]{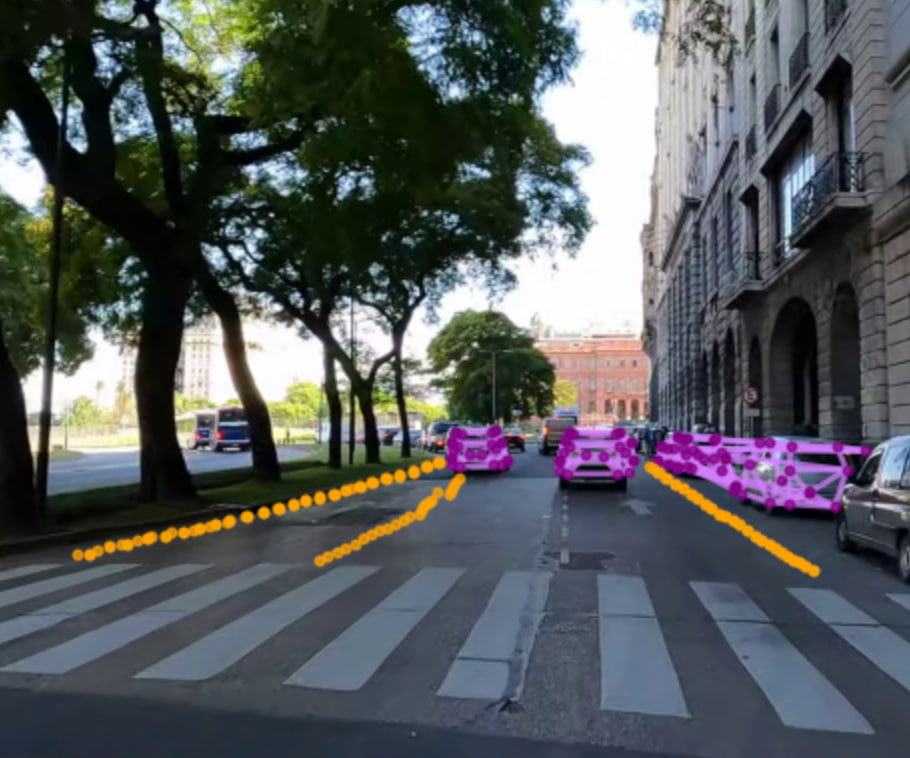}
        \label{fig:fig1}
    \end{subfigure}%
    \hspace{0.01\textwidth}
    \begin{subfigure}[b]{0.45\textwidth}
        \centering
        \includegraphics[height=5cm, width=\linewidth]{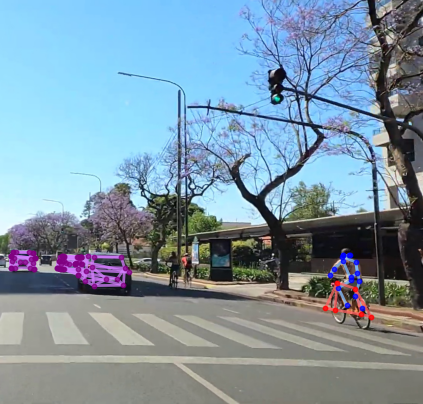}
        \label{fig:fig2}
    \end{subfigure}

    \begin{subfigure}[b]{0.45\textwidth}
        \centering
        \includegraphics[height=5cm, width=\linewidth]{images/dark.png}
        \label{fig:fig3}
    \end{subfigure}%
    \hspace{0.01\textwidth} 
    \begin{subfigure}[b]{0.45\textwidth}
        \centering
        \includegraphics[height=5cm, width=\linewidth]{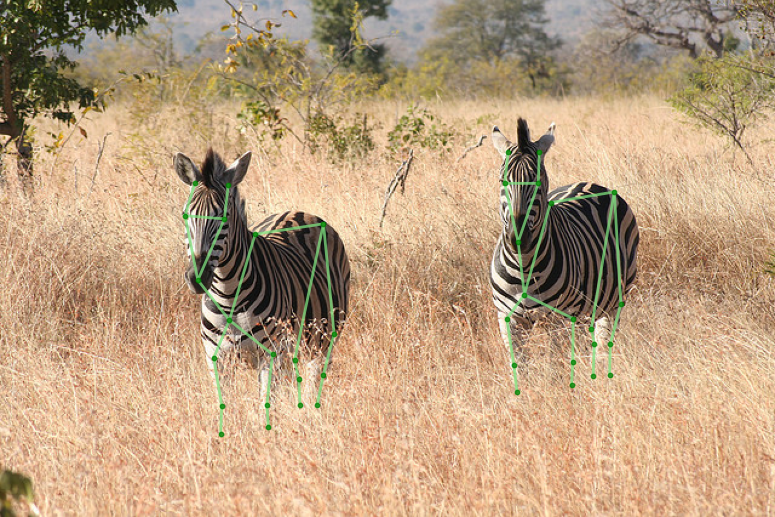}
        \label{fig:fig4}
    \end{subfigure}

    \caption{Experimental results on the OpenDV \cite{yang2024generalized} dataset (first two images) and the MS COCO validation set \cite{lin2014microsoft} (last two images). As shown, our model successfully detects pedestrians (blue), animals (green), cars (purple), bicycles (red), and lanes (orange) in real-world scenarios. More examples are available in our supplementary material.}
    \label{fig:four_figures}
\end{figure*}



\textbf{Feature Fusion Effect.} This study examines the influence of the Feature Pyramid Network (FPN) \cite{lin2017feature} on multi-category skeleton detection. As illustrated in \Cref{fig:overview}, hierarchical fusion of FPN is particularly beneficial for detecting objects of varying scales and complexities.

As shown in \Cref{table:backbone_comparison}, FPN consistently improves performance across different backbones. For the ConvNeXt backbone, integrating FPN increased the mean score from 59.3\% to 64.1\%, demonstrating a substantial performance boost. Similarly, in the ClipConvNeXt backbone, FPN improved the mean score from 63.0\% to 64.6\%. In the Swin-L backbone, the impact of FPN was less pronounced, with the mean score slightly decreasing from 61.2\% to 61.0\%. This suggests that while FPN effectively enhances feature fusion, its effectiveness may vary based on the backbone architecture. Nevertheless, across most settings, FPN contributes positively by enriching feature representations, ultimately improving multi-category skeleton detection performance.

\textbf{Transformer Block.}  
To enhance multi-task performance, we add a task-specific transformer block per category (Swin blocks). This allows the network to prioritize task-relevant features while sharing a backbone, balancing generalization and specialization. As shown in \Cref{table:backbone_comparison}, the task specific transformer block significantly boosts performance across tasks. Adding these layers improved the mean score from 64.1\% to 66.4\% on ConvNeXt, from 64.6\% to 65.5\% on ClipConvNeXt, and from 61.0\% to 65.4\% on Swin-L. Notably, animal detection saw the highest improvement, increasing from 53.6\% to 55.88\% on ConvNeXt, from 52.21\% to 53.51\% on ClipConvNeXt, and from 50.37\% to 53.63\% on Swin-L. While this approach increases model complexity, it substantially enhances feature specificity and overall detection accuracy.


\textbf{Combination of Methods.}  
As demonstrated in \Cref{table:backbone_comparison}, the combined application of Mosaic augmentation, FPN, and attention heads consistently improves performance across all backbones. The most significant gains were observed on the ConvNeXt backbone, where the mean score increased from 59.3\% to 66.4\%, followed by ClipConvNeXt, which improved from 63.0\% to 65.5\%, and Swin-L, which saw an increase from 61.2\% to 65.4\%.  

Notably, the animal detection task, which suffers from data imbalance, saw a remarkable improvement in AP across all backbones, with the highest gain in Swin-L, increasing from 41.0\% to 53.6\%. This highlights the importance of feature fusion, data augmentation, and task-specific attention in overcoming data scarcity challenges and improving multi-category skeleton detection.

The impact of backbone architectures on lane detection with a multi-task network is further detailed in \Cref{tab:Openlane_2}, categorized by various challenging scenarios. It can be observed that the combination of Mosaic augmentation, FPN, and task specific transformer blocks addresses most of the challenging scenarios across all backbones, with the most significant improvement observed with ConvNeXtv2, which increases the $F_1$ score for the Up\&Down cases by 15.3 points. In \Cref{fig:four_figures}, we present the results of our model tested on unseen samples from the OpenDV dataset \cite{yang2024generalized}, which consists of YouTube videos, as well as additional samples from the MS COCO dataset \cite{lin2014microsoft}. 

\color{black}



\begin{figure*}[t]
  \centering
  \includegraphics[width=\textwidth]{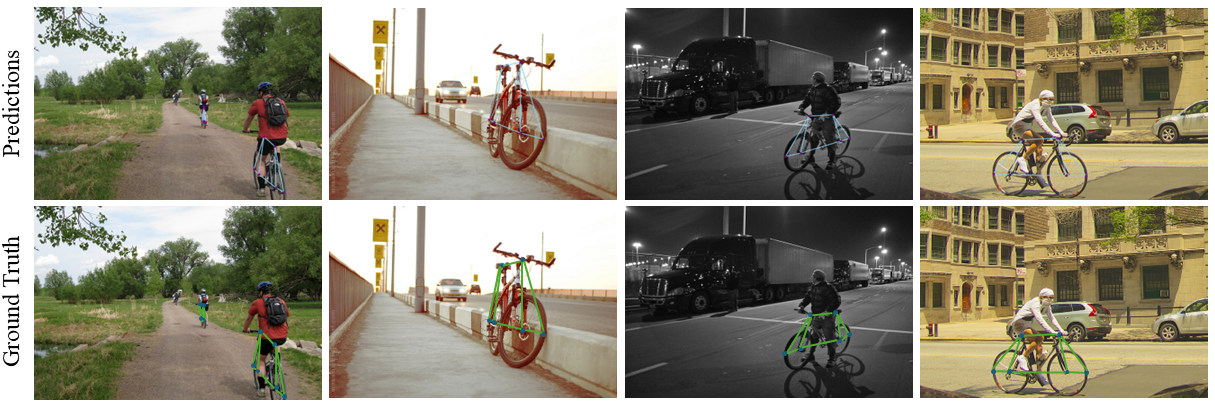}
  \caption{Qualitative results of ConvNeXtv2 on bicycle detection in different lighting conditions.
  }
  \label{fig:bicycle}
\end{figure*}

\begin{table}[t]
    \renewcommand{\arraystretch}{1.2} 
    \caption{Performance is quantified using average precision (AP) computed over IoU\@0.50-0.95. The table illustrates the impact of pretraining on model performance. A \cmark in the "baseline" column denotes that the backbone is employed without incorporating feature pyramid networks (FPN), transformer blocks, or multi-task pretraining; and \xmark indicates that these enhancements are applied.}
    \label{table:backbone_comparison_bicycle}
    \small
    \centering
    \begin{adjustbox}{max width=\textwidth}
    \begin{tabular}{lcccc}  
        \toprule
        \textbf{Backbone} & 
        \textbf{\begin{tabular}{@{}c@{}}Frozen\\Backbone\end{tabular}} & \textbf{Baseline} & \textbf{\begin{tabular}{@{}c@{}}Bicycle\\(AP\%)\end{tabular}} \\ 
        \midrule
        
        \multirow{3}{*}{ConvNeXtv2} 
            & \xmark & \cmark & 73.4 \\ 
            & \cmark & \xmark & 48.5 \\ 
            & \xmark & \xmark & \textbf{79.3} \\ 
        \midrule
        
        \multirow{3}{*}{ClipConvNeXt} 
            & \xmark & \cmark & 69.4 \\
            & \cmark & \xmark & 50.3 \\
            & \xmark & \xmark & \textbf{78.0} \\ 
        \midrule
        
        \multirow{3}{*}{Swin L} 
            & \xmark & \cmark & 71.6 \\ 
            & \cmark & \xmark & 53.1 \\ 
            & \xmark & \xmark & \textbf{80.0} \\ 

        \bottomrule
    \end{tabular}
    \end{adjustbox}
\end{table} 
\subsection{Generalization to New Categories: A Case Study on Bicycles.} This section examines how pre-training on multiple categories affects skeleton detection for an unseen category, bicycles. Models are initially trained on skeleton detection tasks across pedestrians, animals, cars, and lanes.  We then evaluate their generalization by training OpenPifPaf with three different backbone architectures: Swin L, ConvNeXtV2, and ClipConvNeXt, using the bicycle skeletons subset of the MS COCO dataset.

For each backbone, we investigate two variants: (1) a variant where only the bicycle-specific transformer block is trained, while the remaining components remain frozen (attentive probing), and (2) a variant where the entire network is fine-tuned. Both variants include a feature pyramid network (FPN) and a transformer block, and they have been pre-trained on four tasks, as discussed in the previous section, using mosaic augmentation. Additionally, we compare these models against a baseline variant that excludes both the FPN and transformer block and has not undergone multi-task pre-training. This comparison, presented in \Cref{table:backbone_comparison_bicycle}, evaluates the transferability of learned representations to the novel task of bicycle keypoint detection.

Table \Cref{table:backbone_comparison_bicycle} presents the results for the various training settings. The table demonstrates that models pre-trained on four tasks, when fine-tuned on the bicycle dataset, outperform the baseline by a significant margin—achieving improvements of 5.9, 8.6, and 8.4 AP points for ConvNeXtv2, ClipConvNeXt, and Swin-L, respectively. In contrast, the attentive probing strategy—training only the transformer block while keeping other components frozen—resulted in inferior performance relative to both the baseline and full fine-tuning approaches. Notably, freezing components preserves performance on other tasks. These results indicate that the proposed framework effectively generalizes to novel categories for keypoint detection. Moreover, the consistent improvements across all three backbones suggest that the benefits of the proposed modules and multi-task pre-training are not confined to a specific architecture. Future work should explore alternative probing strategies to further close the performance gap observed when the network remains partially frozen.

\color{black}

\section{Implementation Details}
\label{sec:rationale}

\subsection{Lane Detection Training}
The number of keypoints used to model the lane skeleton was chosen as 24, as detailed in the supplementary. We use standard building blocks of ShuffleNetV2 \cite{ma2018shufflenet} to construct our base network for both of our small-sized models (denoted as 'S') and medium-sized models (denoted as 'M'). The base networks have their input max-pooling operation removed as it destroys spatial information \cite{kreiss2021openpifpaf}. A $4\times4$ window is used as a mask for localization and scale components of fields, resulting in a stride of 16 coarse confidence maps with respect to the original images. Standard data augmentation was performed during the training phase, including random horizontal flipping, random rescaling with a rescaling factor $r \in [0.5, 2.0]$, random cropping, and color jittering. Models are trained from pretrained ShuffleNetV2K16/K30 \cite{kreiss2021openpifpaf}, with an initial learning rate of 1.5$e^{-4}$. SGD optimizer \cite{bottou2010large} with Nesterov momentum \cite{nesterov1983method} of 0.95 was used for training, with a batch size of 16 per GPU and weight decay of $10^{-5}$. For the OpenLane dataset, which is larger, we trained for 50 epochs and decayed the learning rate to 3$e^{-5}$ at 30 epochs, and for the CULane dataset, we trained for 150 epochs with a decayed learning rate of 3$e^{-5}$ from 80 epochs with 2 GPUs. All ground truth annotations were processed as described in \Cref{sec:gt} before feeding into the model. During inference, the images were rescaled to a long edge of 621 pixels, keeping the same aspect ratio as the original images. Training and testing were performed on Tesla-V100 GPUs.

\subsection{Bicycle Detection Training}
As described earlier, we annotated six distinct keypoints on bicycles for skeleton detection. For single-task training, the backbones were trained for 100 epochs using a single GPU. All other hyperparameter settings were kept consistent with those used for lane detection training. Figure \Cref{fig:bicycle} presents a qualitative comparison between the ground truth and the predictions produced by ConvNeXtv2. The quantitative results are presented in \Cref{table:backbone_comparison_bicycle}.

\subsection{Multi-Class Training}
This section provides the implementation details for the experiments discussed in \Cref{sec:multi_task_sec}. The multi-class training process is divided into two stages:

\begin{itemize}
    \item \textbf{Single-Class Training:} The model is initially trained with a single head for human keypoint detection on the MS COCO dataset. Training is conducted for 250 epochs using a batch size of 64, an initial learning rate of 1$e^{-3}$, and a linear learning rate scheduler.
    
    \item \textbf{Multi-Class Training:} In this stage, the weights from the first stage (for the backbone and human skeleton detection head) are used as initialization. Additional heads for animals, vehicles, and lanes are randomly initialized. The model is trained for 200 epochs in this stage. For all backbones, the batch size remains 64. The initial learning rate is set to 5$e^{-4}$ for ConvNeXt and ClipConvNeXt backbones, and 5$e^{-5}$ for the SwinL backbone.
\end{itemize}

The optimizer used in both stages is AdamW \cite{loshchilov2017fixing}. During multi-class training, to address data imbalance, training for each epoch ends when all samples from the smallest dataset have been used. For example, if the human dataset contains 10,000 samples and the animal dataset contains 4,000 samples, each epoch includes 4,000 samples from both datasets, assuming a batch weighting of 0.5 for each. For ConvNeXt and ClipConvNeXt backbones, batch weights are distributed equally across all four classes, whereas for the SwinL backbone, the weight for humans is set to half that of the other classes.

\section{Conclusion}


In this work, we introduced a unified bottom-up framework for multi-category skeleton detection, efficiently handling diverse object classes such as pedestrians, animals, cars, bicycles, and lanes. Our approach requires no auxiliary inputs beyond standard images while maintaining strong performance across tasks. Notably, our skeleton-based lane detection method achieves state-of-the-art results, highlighting the effectiveness of skeleton representations for this task. Additionally, we introduced a new set of bicycle skeleton annotations and demonstrated that multi-task pre-training enhances performance on this unaddressed category. Leveraging shared representations across object classes showcases the potential of skeleton-based models for transfer learning. Our findings emphasize the effectiveness of keypoint representations in autonomous driving and related applications, advancing multi-category skeleton detection.

\section{Acknowledgments}
The authors gratefully acknowledge the financial support provided by the Swiss National Science Foundation (SNSF) through grant no. 10003100.

{
    \small
    \printbibliography
}

\end{document}